\documentclass[11pt,a4paper]{article}
\usepackage[hyperref]{emnlp2018}
\usepackage{times,comment,graphicx,lipsum}
\usepackage{latexsym,amsmath,amssymb}
\usepackage{xcolor,soul}
\sethlcolor{lightgray}
\usepackage{url,xfrac,multirow}

\aclfinalcopy % Uncomment this line for the final submission
\def\aclpaperid{1465} %  Enter the acl Paper ID here

\title{Disfluency Detection using Auto-Correlational Neural Networks}

\author{Paria Jamshid Lou \\
  Macquarie University \\
  Sydney, Australia \\
  {\tt paria.jamshid-lou@hdr.mq.edu.au } \\\And
  Peter Anderson \\
  Macquarie University \\
  Sydney, Australia\\
  {\tt p.anderson@mq.edu.au } \\\AND
  Mark Johnson \\
  Macquarie University \\
  Sydney, Australia\\
  {\tt mark.johnson@mq.edu.au } \\}

\date{}

\begin{document}
\maketitle
\begin{abstract}
In recent years, the natural language processing community has moved away from task-specific feature engineering, i.e., researchers discovering ad-hoc feature representations for various tasks, in favor of general-purpose methods that learn the input representation by themselves. However, state-of-the-art approaches to disfluency detection in spontaneous speech transcripts currently still depend on an array of hand-crafted features, and other representations derived from the output of pre-existing systems such as language models or dependency parsers. As an alternative, this paper proposes a simple yet effective model for automatic disfluency detection, called an auto-correlational neural network (ACNN). The model uses a convolutional neural network (CNN) and augments it with a new auto-correlation operator at the lowest layer that can capture the kinds of ``rough copy'' dependencies that are characteristic of repair disfluencies in speech. In experiments, the ACNN model outperforms the baseline CNN on a disfluency detection task with a 5\% increase in f-score, which is close to the previous best result on this task\footnote{\url{https://github.com/pariajm/deep-disfluency-detector}}.

\end{abstract}

\section{Introduction}
\label{sec:intro}

Disfluency informally refers to any interruptions in the normal flow of speech, including false starts, corrections, repetitions and filled pauses. \citet{shri:94} defines three distinct parts of a speech disfluency, referred to as the \textit{reparandum}, \textit{interregnum} and \textit{repair}. As illustrated in Example \ref{ex:1}, the reparandum \textit{to Boston} is the part of the utterance that is replaced, the interregnum \textit{uh, I mean} (which consists of a filled pause \textit{uh} and a discouse marker \textit{I mean}) is an optional part of a disfluent structure, and the repair \textit{to Denver} replaces the reparandum. The fluent version is obtained by removing reparandum and interregnum words although disfluency detection models mainly deal with identifying and removing reparanda. The reason is that filled pauses and discourse markers belong to a closed set of words and phrases and are trivial to detect~\citep{john:04}.

\begin{equation}  \label{ex:1}
\begin{array}{l}
\mbox{\it I want a flight~}\overbrace{\mbox{\it\strut to Boston,}}^{\mbox{\tiny reparandum}}\\
\strut\hspace{-0.5cm}\underbrace{\mbox{\it{}uh, I mean\strut}}_{\mbox{\tiny interregnum}} \underbrace{\mbox{\it\strut to Denver~}}_{\mbox{\tiny repair}} \mbox{\it on~Friday}
\end{array}
\end{equation}

In disfluent structures, the repair (e.g., \textit{to Denver}) frequently seems to be a ``rough copy" of the reparandum (e.g., \textit{to Boston}). In other words, they incorporate the same or very similar words in roughly the same word order. In the Switchboard training set~\cite{godfrey:93}, over $60\%$ of the words in the reparandum are exact copies of words in the repair. Thus, this similarity is strong evidence of a disfluency that can help the model detect reparanda~\citep{char:01, john:04}. As a result, models which are able to detect ``rough copies" are likely to perform well on this task.

Currently, state-of-the-art approaches to disfluency detection depend heavily on hand-crafted pattern match features, specifically designed to find such ``rough copies''~\citep{zay:16, jam:17}. In contrast to many other sequence tagging tasks~\cite{plank2016multilingual,yu2017general}, ``vanilla" convolutional neural networks (CNNs) and long short-term memory (LSTM) models operating only on words or characters are surprisingly poor at disfluency detection~\cite{zay:16}. As such, the task of disfluency detection sits in opposition to the ongoing trend in NLP away from task-specific feature engineering --- i.e., researchers discovering ad-hoc feature representations for various tasks --- in favor of general-purpose methods that learn the input representation by themselves~\cite{collobert2008unified}.

In this paper, we hypothesize that LSTMs and CNNs cannot easily learn ``rough copy'' dependencies. We address this problem in the context of a CNN by introducing a novel auto-correlation operator. The resulting model, called an \textit{auto-correlational neural network (ACNN)}, is a generalization of a CNN with an auto-correlation operator at the lowest layer. Evaluating the ACNN in the context of disfluency detection, we show that introducing the auto-correlation operator increases f-score by 5\% over a baseline CNN. Furthermore, the ACNN --- operating only on word inputs --- achieves results which are competitive with much more complex approaches relying on hand-crafted features and outputs from pre-existing systems such as language models or dependency parsers. In summary, the main contributions of this paper are:

\begin{itemize}
	\item We introduce the auto-correlational neural network (ACNN), a generalization of a CNN incorporating auto-correlation operations,
	\item In the context of disfluency detection, we show that the ACNN captures important properties of speech repairs including ``rough copy'' dependencies, and
	\item Using the ACNN, we achieve competitive results for disfluency detection without relying on any hand-crafted features or other representations derived from the output of pre-existing systems.
\end{itemize}

\section{Related Work} \label{sec:2}
Approaches to disfluency detection task fall into three main categories: noisy channel models, parsing-based approaches and sequence tagging approaches. Noisy channel models (NCMs)~\citep{john:04, john:04a} use complex tree adjoining grammar (TAG)~\citep{shie:90} based channel models to find the ``rough copy'' dependencies between words. The channel model uses the similarity between the reparandum and the repair to allocate higher probabilities to exact copy reparandum words. Using the probabilities of TAG channel model and a bigram language model (LM) derived from training data, the NCM generates $n$-best disfluency analyses for each sentence at test time. The analyses are then reranked using a language model which is sensitive to the global properties of the sentence, such as a syntactic parser based LM~\citep{john:04, john:04a}. Some works have shown that rescoring the $n$-best analyses with external $n$-gram~\citep{zwa:11} and deep learning LMs~\citep{jam:17} trained on large speech and non-speech corpora, and using the LM scores along with other features (i.e. pattern match and NCM ones) into a MaxEnt reranker~\citep{john:04a} improves the performance of the baseline NCM, although this creates complex runtime dependencies. 

Parsing-based approaches detect disfluencies while simultaneously identifying the syntactic structure of the sentence. Typically, this is achieved by augmenting a transition-based dependency parser with a new action to detect and remove the disfluent parts of the sentence and their dependencies from the stack~\citep{ras:13, hon:14, yoshi:16}. Joint parsing and disfluency detection can compare favorably to pipelined approaches, but requires large annotated tree-banks containing both disfluent and syntatic structures for training. 

Our proposed approach, based on an auto-correlational neural network (ACNN), belongs to the class of sequence tagging approaches. These approaches use classification techniques such as conditional random fields~\citep{liu:06, ost:13, zay:14, fer:15}, hidden Markov models~\citep{liu:06, schul:10} and deep learning based models~\citep{hough:15, zay:16} to label individual words as fluent or disfluent. In much of the previous work on sequence tagging approaches, improved performance has been gained by proposing increasingly complicated labeling schemes. In this case, a model with begin-inside-outside (BIO) style states which labels words as being inside or outside of edit region\footnote{For state labels, edit corresponds to reparandum.} is usually used as the baseline sequence tagging model. Then in order to come up with different pattern matching lexical cues for repetition and correction disfluencies, they extend the baseline state space with new explicit repair states to consider the words at repair region, in addition to edit region~\citep{ost:13, zay:14, zay:16}. A model which uses such labeling scheme may generate illegal label sequences at test time. As a solution, integer linear programming (ILP) constraints are applied to the output of classifier to avoid inconsistencies between neighboring labels~\citep{geo:09, geo:10, zay:16}. This contrasts with our more straightforward approach, which directly labels words as being fluent or disfluent, and does not require any post-processing or annotation modifications. 

The most similar work to ours is recent work by \citet{zay:16} that investigated the performance of a bidirectional long-short term memory network (BLSTM) for disfluency detection. \citet{zay:16} reported that a BLSTM operating only on words underperformed the same model augmented with hand-crafted pattern match features and POS tags by 7\% in terms of f-score. In addition to lexically grounded features, some works incorporate prosodic information extracted from speech~\citep{kahn:05, fer:15, trang:18}. In this work, our primary motivation is to rectify the architectural limitations that prevent deep neural networks from automatically learning appropriate features from words alone. Therefore, our proposed model eschews manually engineered features and other representations derived from dependency parsers, language models or tree adjoining grammar transducers that are used to find ``rough copy'' dependencies. Instead, we aim to capture these kinds of dependencies automatically.

\section{Convolutional and Auto-Correlational Networks} \label{sec:3.1}

\newcommand{\bx}{{\boldsymbol{x}}}
\newcommand{\by}{{\boldsymbol{y}}}
\newcommand{\bb}{{\boldsymbol{b}}}
\newcommand{\bu}{{\boldsymbol{u}}}
\newcommand{\bv}{{\boldsymbol{v}}}

In this section, we introduce our proposed auto-correlation operator and the resulting auto-correlational neural network (ACNN) which is the focus of this work. 

A convolutional or auto-correlational network computes a series
of $h$ feature representations ${X}^{(0)}, {X}^{(1)}, \ldots, {X}^{(h)}$, where 
${X}^{(0)}$ is the input data, ${X}^{(h)}$ is the final (output) representation,
and each non-input representation ${X}^{(k)}$ for $k>0$, is computed from 
the preceding representation ${X}^{(k-1)}$ using a convolution or
auto-correlation operation followed by an element-wise non-linear function.

Restricting our focus to convolutions in one dimension, as used in the context of 
language processing, each representation ${X}^{(k)}$ is a matrix of size $(n,m_k)$,
where $n$ is the number of words in the input and $m_k$ is the feature dimension of representation $k$, or equivalently it can
be viewed as a sequence of $n$ row vectors ${X}^{(k)} = (\bx^{(k)}_1, \ldots, \bx^{(k)}_n)$, where $\bx^{(k)}_t$ is the row vector of length $m_k$ that represents the $t$th word at level $k$.

Consistent with the second interpretation, the input representation ${X}^{(0)}=(\bx^{(0)}_1,\ldots,\bx^{(0)}_n)$
is a sequence of word embeddings, where $m_0$ is the length of
the embedding vector and $\bx^{(0)}_t$ is the word embedding for
the $t$th word.  

Each non-input representation ${X}^{(k)}, k>0$ is formed by column-wise stacking
the output of one or more convolution or auto-correlation operations
applied to the preceding representation, and then applying an element-wise non-linear function. Formally, we define:
\begin{align}
{Y}^{(k)}=&\left({F}^{(k,1)}({X}^{(k-1)}); \ldots; {F}^{(k,m_k)}({X}^{(k-1)})\right) \nonumber \\
{X}^{(k)}=&{N}^{(k)}({Y}^{(k)})
\end{align}
\noindent
where ${F}^{(k,u)}$ is the $u$th operator applied at layer $k$, and ${N}^{(k)}$ is the non-linear operation applied at layer $k$. Each operator ${F}^{(k,u)}$ (either convolution or auto-correlation) is a function from ${X}^{(k-1)}$, which is a matrix of size $(n,m_{k-1})$, to a vector of length $n$. A network that employs only convolution operators is a convolutional neural network (CNN). We call a network that utilizes a mixture of convolution and auto-correlation operators an \emph{auto-correlational neural network} (ACNN). In our networks, the non-linear operation ${N}^{(k)}$ is always element-wise $\mathop{ReLU}$, except for the last layer, which uses a $\mathop{softmax}$ non-linearity.

\subsection{Convolution Operator} \label{sec:3.2}
A one-dimensional convolution operation maps an input matrix ${X} = (\bx_1,\ldots,\bx_n)$, where
each $\bx_t$ is a row vector of length $m$, to an output vector $\by$ of length $n$. The convolution operation is defined by a convolutional kernel ${A}$, which is applied to a window of words to produce a new output representation, and kernel width parameters $\ell$ and $r$, which define the number of words to the left and right of the target word included in the convolutional window. For example, assuming appropriate input padding where necessary, element $y_t$ in the output vector $\by$ is computed as: 
\begin{eqnarray}
y_t & = & A \cdot X_{i:j} + \bb
\end{eqnarray}
where
\begin{description}
	\item[${A}$] is a learned convolutional kernel of dimension $(\ell+r, m)$,
	\item[$X_{i:j}$] is the sub-matrix formed by selecting rows $i$ to $j$ from matrix $X$,
	\item[$\cdot$] is the dot product (a sum over elementwise multiplications),
	\item[$i,j$] are given by $i=t-\ell$ and $j=t+r$, indicating the left and right extremities of the convolutional window effecting element $y_t$, 
	\item[$\ell>0$] is the left kernel width, and
	\item[$r>0$] is right kernel width.
	\item[$\bb$] is a learned bias vector of dimension $n$,
\end{description}

\begin{figure} 
	\centering
	\includegraphics[width=0.38\textwidth]{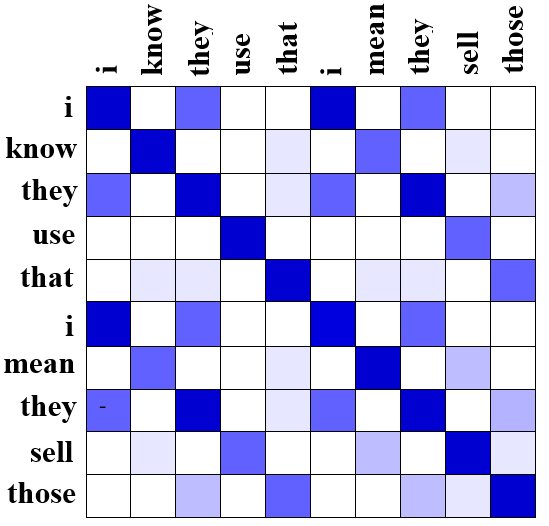}
	\caption{Cosine similarity between word embedding vectors learned by the ACNN model for the sentence ``I know \hl{they use that} I mean they sell those'' (with disfluent words highlighted). In the figure, darker shades denote higher cosine values. ``Rough copies'' are clearly indicated by darkly shaded diagonals, which can be detected by our proposed auto-correlation operator. }
	\label{fig:01}
\end{figure}

\begin{figure*}[ht] 
\centering
\includegraphics[width=0.77\textwidth]{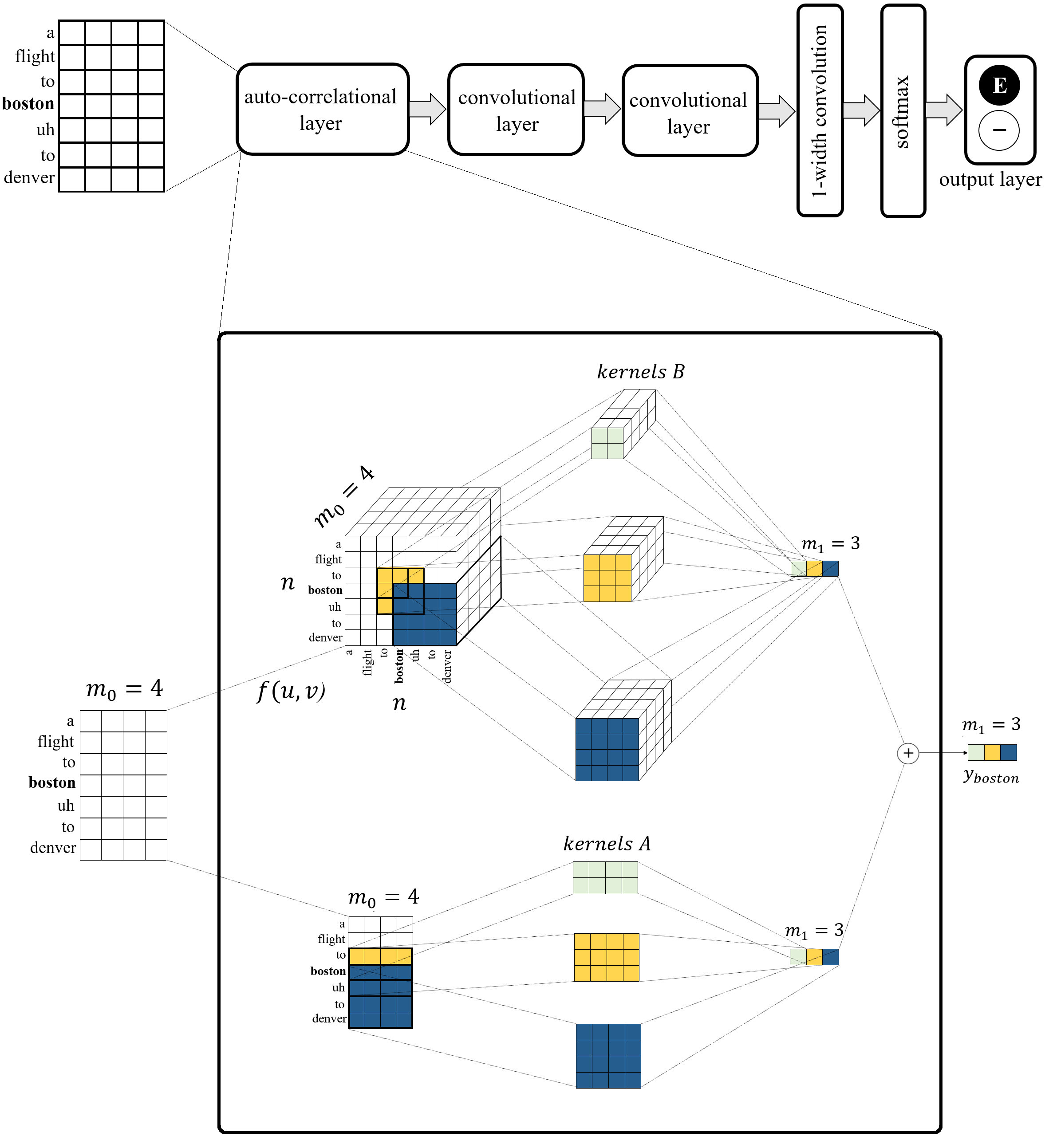}
\caption{ACNN overview for labeling the target word ``boston''. A patch of words is fed into an auto-correlational layer. At inset bottom, the given patch of words is convolved with 2D kernels $A$ of different sizes. At inset top, an auto-correlated tensor of size $(n,n,m_0)$ is constructed by comparing each input vector $\bu = \bx_t$ with the input vector $\bv = \bx_{t'}$ using a binary function $f(\bu,\bv)$. The auto-correlated tensor is convolved with 3D kernels $B$ of different sizes. Each kernel group $A$ and $B$ outputs a matrix of size $(n,m_1)$ (here, we depict only the row vector relating to the target word ``boston''). These outputs are added element-wise to produce the feature representation that is passed to further convolutional layers, followed by a softmax layer. ``E'' = disfluent, ``$\_$'' = fluent and $m_0$ = embedding size. }
\label{fig:02}
\end{figure*}

\subsection{Auto-Correlation Operator} \label{sec:3.3}
The auto-correlational operator is a generalisation of the convolution operator: 
\begin{eqnarray}
\label{eq:ac}
y_t & = & A \cdot X_{i:j} + B \cdot \hat{X}_{i:j,i:j} + \bb
\end{eqnarray}
where $y_t$, $A$, $X$, $\bb$, $i$ and $j$ are as in the convolution operator, and
\begin{description}
	\item [$\hat{X}$] is a tensor of size $(n,n,m)$ such that each vector $\hat{X}_{i,j,:}$ is given by $f(\bx_i,\bx_j)$,
	\item[$f(\bu,\bv)$] is a binary operation on vectors, such as the Hadamard or element-wise product (i.e., $f(\bu,\bv) = \bu\circ \bv$), and
	\item[$\hat{X}_{i:j,i:j}$] is the sub-tensor formed by selecting indices $i$ to $j$ from the first two dimensions of tensor $\hat{X}$,
	\item[${B}$] is a learned convolutional kernel of dimension $(\ell + r,\ell + r, m)$.
\end{description}
Unlike convolution operations, which are linear, the auto-correlation operator introduces second-order interaction terms through the tensor $\hat{X}$ (since it multiplies the vector representations for each pair of input words). This naturally encodes the similarity between input words when applied at level $k=1$ (or the co-activations of multiple CNN features, if applied at higher levels). As illustrated in Figure~\ref{fig:01}, blocks of similar words are indicative of ``rough copies''. We provide an illustration of the auto-correlation operation in Figure~\ref{fig:02}.

\section{Experiments} \label{sec:4}

\subsection{Switchboard Dataset}
We evaluate the proposed ACNN model for disfluency detection on the Switchboard corpus of conversational speech~\cite{godfrey:93}. Switchboard is the largest available corpus  ($1.2\times10^{6}$ tokens) where disfluencies are annotated according to Shriberg's~\citeyearpar{shri:94} scheme:
\begin{center}
	[ reparandum + \{interregnum\} repair ] 
\end{center}
where (+) is the interruption point marking the end of reparandum and \{\} indicate optional interregnum. We collapse this annotation to a binary classification scheme in which reparanda are labeled as disfluent and all other words as fluent. We disregard interregnum words as they are trivial to detect as discussed in Section \ref{sec:intro}. 

Following Charniak and Johnson~\shortcite{char:01}, we split the Switchboard corpus into training, dev and test set as follows: training data consists of all sw[23]$\ast$.dff files, dev training consists of all sw4[5-9]$\ast$.dff files and test data consists of all sw4[0-1]$\ast$.dff files. We lower-case all text and remove all partial words and punctuations from the training data to make our evaluation both harder and more realistic~\citep {john:04}. Partial words are strong indicators of disfluency; however, speech recognition models never generate them in their outputs.

\subsection{ACNN and CNN Baseline Models}

We investigate two neural network models for disfluency detection; our proposed auto-correlational neural network (ACNN) and a convolutional neural network (CNN) baseline. The CNN baseline contains three convolutional operators (layers), followed by a width-1 convolution and a softmax output layer (to label each input word as either fluent or disfluent). The ACNN has the same general architecture as the baseline, except that we have replaced the first convolutional operator with an auto-correlation operator, as illustrated in Figure~\ref{fig:02}. 

\begin{figure}
 \centering
\includegraphics[width=0.48\textwidth]{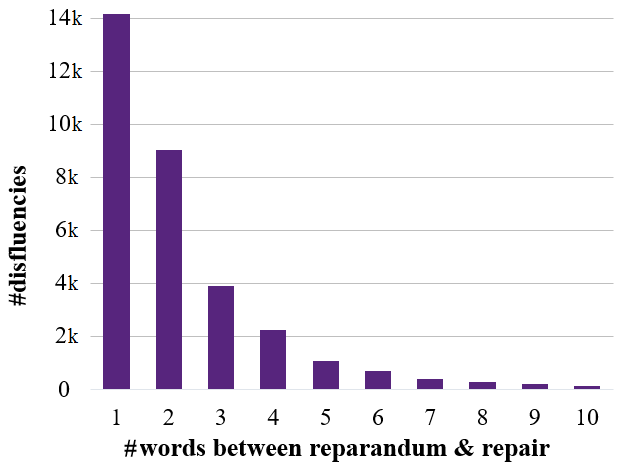}
\caption{Distribution over the number of words in between the reparandum and the interregnum in the Switchboard training set (indicating the distance between ``rough copies''). }
\label{fig:03}
\end{figure}

To ensure that equal effort was applied to the hyperparameter optimization of both models, we use randomized search~\cite{berg:12} to tune the optimization and architecture parameters separately for each model on the dev set, and to find an optimal stopping point for training. This results in different dimensions for each model. As indicated by Table~\ref{tab:01}, the resulting ACNN configuration has far fewer kernels at each layer than the CNN. However, as the auto-correlation kernels contain an additional dimension, both models have a similar number of parameters overall. Therefore, both models should have similar learning capacity except for their architectural differences (which is what we wish to investigate). Finally, we note that the resulting maximum right kernel width $r_1$ in the auto-correlational layer is 6. As illustrated in Figure \ref{fig:03}, this is sufficient to capture almost all the ``rough copies'' in the Switchboard dataset (but could be increased for other datasets). 

\begin{table}[h!]
\begin{center}
\begin{tabular}{|l|c|c|}
\hline \bf Configuration & \bf  CNN & \bf ACNN  \\ \hline
embedding dim & $290$ & $290$  \\ \hline
dropout rate & $0.51$ & $0.53$ \\ \hline
$L_2$ regularizer weight & $0.13$ & $0.23$ \\ \hline
$\#$kernels at each layer & $570$ & $120$ \\ \hline 
$\#$kernel sizes at each layer & $3$ & $2$ \\ \hline 
$\#$words at left context $\ell_1$ & [0,1,4] & [5,3] \\ \hline
$\#$words at left context $\ell_2$ & [1,2,3] & [4,2] \\ \hline
$\#$words at left context $\ell_3$ & [0,1,2] & [3,2] \\ \hline
$\#$words at right context $r_1$ & [1,1,4] & [6,3] \\ \hline
$\#$words at right context $r_2$ & [1,2,4] & [5,3] \\ \hline
$\#$words at right context $r_3$ & [1,2,3] & [4,2] \\ \hline
$\#$parameters & 4.9M & 4.9M \\ \hline
\end{tabular}
\end{center}
\caption{Configuration of the CNN and ACNN models, where $\ell_k$ refers to the left kernel width at layer $k$, and $r_k$ refers to the right kernel width at layer $k$. Both models have a similar total number of parameters.}\label{tab:01} 
\end{table}

For the ACNN, we considered a range of possible binary functions $f(\bu, \bv)$ to compare the input vector $\bu = \bx_t$ with the input vector $\bv = \bx_{t'}$ in the auto-correlational layer. However, in initial experiments we found that the Hadamard or element-wise product (i.e. $f(\bu,\bv)=\bu\circ \bv$) achieved the best results. We also considered concatenating the outputs of kernels $A$ and $B$ in Equation \ref{eq:ac}, but we found that element-wise addition produced slightly better results on the dev set. 

\subsubsection{Implementation Details}

In both models, we use $\mathop{ReLU}$ for the non-linear operation, all stride sizes are one word and there are no pooling operations. We randomly initialize the word embeddings and all weights of the model from a uniform distribution. The bias terms are initialized to be $1$. To reduce overfitting, we apply dropout~\cite{hint:12a} to the input word embeddings and $L_2$ regularization to the weights of the width-1 convolutional layer. For parameter optimization, we use the Adam optimizer~\citep{king:14} with a mini-batch size of $25$ and an initial learning rate of $0.001$.

\section{Results}

As in previous work~\citep{john:04}, we evaluate our model using precision, recall and f-score, where true positives are the words in the edit region (i.e., the reparandum words). 
As~\citet{char:01} observed, only $6\%$ of words in the Switchboard corpus are disfluent, so accuracy is not a good measure of system performance. F-score, on the other hand, focuses more on detecting ``edited'' words, so it is more appropriate for highly skewed data. 

Table~\ref{tab:02} compares the dev set performance of the ACNN model against our baseline CNN, as well as the LSTM and BLSTM models proposed by \citet{zay:16} operating only on word inputs (i.e., without any disfluency pattern-match features). Our baseline CNN outperforms both the LSTM and the BLSTM, while the ACNN model clearly outperforms the baseline CNN, with a further 5\% increase in f-score. In particular, the ACNN noticeably improves recall without degrading precision.

\begin{table}[h!]
	\begin{center}
		\begin{tabular}{|p{3cm}|c|c|c|}
			\hline \bf model & \bf  P & \bf R & \bf  F \\ \hline
			BLSTM\small{~(words)}$^*$ & $87.8$ & $71.1$ & $78.6$ \\ \hline
			LSTM\small{~(words)}$^*$ & $87.6$ & $71.4$ & $78.7$ \\ \hline
			CNN &  $89.4$ & $74.6$ &  $81.3$ \\ \hline
			ACNN & $\bf90.0$ & $\bf82.8$ &  $\bf86.2$ \\
			\hline
		\end{tabular}
	\end{center}
	\caption{Precision (P), recall (R) and f-score (F) on the dev set for the BLSTM and LSTM models using words alone from $^*$\citet{zay:16}, as well as our baseline CNN and ACNN model. }\label{tab:02} 
\end{table}

To further investigate the differences between the two CNN-based models, we randomly select $100$ sentences containing disfluencies from the Switchboard dev set and categorize them according to Shriberg's~\citeyearpar{shri:94} typology of speech repair disfluencies.  \emph{Repetitions} are repairs where the reparandum and repair portions of the disfluency are identical, while \emph{corrections} are where the reparandum and repairs differ (so corrections are much harder to detect).  \emph{Restarts} are where the speaker abandons a sentence prefix, and starts a fresh sentence.
As Table~\ref{tab:03} shows, the ACNN model is better at detecting repetition and correction disfluencies than the CNN, especially for the more challenging correction disfluencies.  On the other hand, the ACNN is no better than the baseline at detecting restarts, probably because the restart typically does not involve a rough copy dependency. Luckily restarts are much rarer than repetition and correction disfluencies.

\begin{table}[h!]
	\begin{center}
		\begin{tabular}{|l|c|c|c|c|}
			\hline \bf model & \bf  Rep. & \bf Cor. & \bf  Res. & \bf All \\ \hline
			CNN &  $93.3$ & $66.0$ &  $57.1$ & $80.4$ \\ \hline
			ACNN & $97.5$ & $80.0$ &  $57.1$ & $88.9$ \\
			\hline
		\end{tabular}
	\end{center}
	\caption{F-scores for different types of disfluencies on a subset of the Switchboard dev set containing $140$ disfluent structures --- including $85$ repetitions (Rep.), $51$ corrections (Cor.) and $4$ restarts (Res.). }\label{tab:03} 
\end{table}

We also repeated the analysis of~\citep{zay:14} on the dev data, so we can compare our models to their extended BLSTM model with a 17-state CRF output and hand-crafted features, including partial-word and POS tag features that enable it to capture some ``rough copy'' dependencies.  As expected, the ACNN outperforms both the CNN and the extended BLSTM model, especially in the ``Other'' category that involve the non-repetition dependencies.

\begin{table}[h!]
	\begin{center}
		\begin{tabular}{|l|c|c|c|c|}
			\hline \bf model & \bf  Rep. & \bf Other & \bf  Either \\ \hline
			CNN & $92.2$  & $66.7$ &  $81.3$ \\ \hline 
			BLSTM (17 states)$^*$ & $94.1$ & $66.7$  & $85.8$ \\ \hline
			ACNN & $96.6$ & $73.3$ &  $86.2$ \\
			\hline
		\end{tabular}
	\end{center}
	\caption{F-scores for different types of disfluencies for the CNN, ACNN and BLSTM (17 states)~$^*$\citep{zay:16} using the Switchboard dev set. }\label{tab:04} 
\end{table}

Finally, we compare the ACNN model to state-of-the-art methods from the literature, evaluated on the Switchboard test set. Table~\ref{tab:05} shows that the ACNN model is competitive with recent models from the literature.  The three models that score more highly than the ACNN all rely on hand-crafted features, additional information sources such as partial-word features (which would not be available in a realistic ASR application), or external resources such as dependency parsers and language models. 
The ACNN, on the other hand, only uses whole-word inputs and learns the ``rough copy'' dependencies between words without requiring any manual feature engineering.

\begin{table}[h]
	\begin{center}
		\begin{tabular}{|p{3.73cm}|c|c|c|}
			\hline \bf model & \bf P & \bf R & \bf F \\ \hline
			
			Yoshikawa et al.\citeyearpar{yoshi:16}\small{~$\bf\diamond$} & $67.9$ & $57.9$ & $62.5$ \\  
			Georgila et al.~\citeyearpar{geo:10}\small{$~\bf\dagger$}  & $77.4$ & $64.6$ & $70.4$ \\			
			Tran et al.~\citeyearpar{trang:18}\small{$~\bf\otimes\star$}  & - & - & $77.5$ \\			
			Kahn et al.~\citeyearpar{kahn:05}\small{~$\bf\star$} &  - & - &  $78.2$ \\ 
			Johnson et al.~\citeyearpar{john:04}\small {~$\bf\wr$} &  $82.0$ & $77.8$ & $79.7$ \\ 
			Georgila~\citeyearpar{geo:09}\small{$~\bf\dagger$}  & - & - &  $80.1$ \\ 
			Johnson et al.~\citeyearpar{john:04a}\small {~$\bf\dagger\wr$} &  - & - & $81.0$   \\ 
			Rasooli et al.~\citeyearpar{ras:13}\small{~$\bf\diamond$} & $85.1$ & $77.9$ & $81.4$ \\
			Zwarts et al.~\citeyearpar{zwa:11}~\small{$\bf\Join\wr$}  & - & -  &  $83.8$ \\ 
			Qian et al.~\citeyearpar{qian:13}\small{~$\bf\Join$}  & - & -  & $84.1$ \\
			Honnibal et al.~\citeyearpar{hon:14}\small{~$\bf\diamond$} & - & - & $84.1$ \\ \hline
			\bf ACNN & $\bf 89.5$ & $\bf 80.0$ & $\bf 84.5$  \\ \hline
			%Ferguson et al.~\citeyearpar{fer:15}  &  $90.1$  & $80.0$ & $84.8$ \\
			Ferguson et al.~\citeyearpar{fer:15}\small{~$\bf\star$}  &  $90.0$  & $81.2$ & $85.4$ \\
			Zayats et al.~\citeyearpar{zay:16}\small{~$\bf\otimes\dagger$} &  $91.8$ & $80.6$ & $85.9$ \\
			\small {Jamshid Lou et al.~\citeyearpar{jam:17}}\small~{$\bf\Join\wr$} & - & - & $86.8$ \\ \hline

		\end{tabular}
	\end{center}
	\caption{Comparison of the ACNN model to the state-of-the-art methods on the Switchboard test set. The other models listed have used richer inputs and/or rely on the output of other systems, as well as pattern match features, as indicated by the following symbols: $\bf\diamond$ dependency parser, $\bf\dagger$ hand-crafted constraints/rules, $\bf\star$~prosodic cues, $\bf\wr$~tree adjoining grammar transducer, $\bf\Join$~refined/external language models and $\bf\otimes$~partial words. P = precision, R = recall and F = f-score.}\label{tab:05} 
\end{table}

\subsection{Qualitative Analysis} \label{sec:5}
We conduct an error analysis on the Switchboard dev set to characterize the disfluencies that the ACNN model can capture and those which are difficult for the model to detect. In the following examples, the \hl{highlighted} words indicate ground truth disfluency labels and the \underline{underlined} ones are the ACNN predictions.

\begin{enumerate}
\item But \hl{{\underline {\smash {if you let them}}}} yeah if you let them in a million at a time \hl{{\underline{it wouldn't make that}}} you know it wouldn't make that big a bulge in the population \\

\vspace{-0.5cm}

\item \hl{{\underline{\smash{They're handy uh they}}}} they come in handy at the most unusual times \\

\vspace{-0.5cm}

\item My mechanics loved it because \hl{{\underline{it was an old}}} it was a sixty-five buick \\

\vspace{-0.5cm}

\item Well \hl{{\underline{I I I think we did}}} I think we did learn some lessons that \hl{{\underline{we weren't}}} uh we weren't prepared for \\

\vspace{-0.5cm}

\item Uh \hl{{\underline{I have never even}}} I have never even looked at one closely \\

\vspace{-0.5cm}

\item But uh \hl{{\underline{when I was}}} when my kids were young I was teaching at a university  \\

\vspace{-0.5cm}

\item She said she'll never put her child \hl{{\underline{in a in a in a in a}}} in a preschool  \\

\vspace{-0.5cm}

\item Well I think \hl{{\underline{\smash{they're at they're}}}} they've come a long way \\

\vspace{-0.5cm}

\item \hl{{\underline{\smash{I I like a}}}} I saw \hl{{\underline{\smash{the the the}}}} the tapes \hl{{\underline{\smash{that were}}}} that were run of marion berry's drug bust  \\

\vspace{-0.5cm}

\item But \hl{{\underline{\smash{I know that in some}}}} I know in a lot of rural areas they're not that good \\

\end{enumerate}
\vspace{-0.5cm}

According to examples 1-10, the ACNN detects \emph{repetition} (e.g. 1, 5) and \emph{correction} disfluencies (e.g. 3, 6, 10). It also captures complex structures where there are multiple or nested disfluencies (e.g. 2, 8) or stutter-like repetitions (e.g. 4, 7, 9).

\begin{enumerate}
\item [11.] My point was that \hl{there is} for people who don't want to do the military service it would be neat if there were an alternative $\ldots$ \\

\vspace{-0.5cm}

\item [12.] I believe from what I remember of the literature \hl{they gave} uh if you fail I believe they give you one more chance \\

\vspace{-0.5cm}

\item [13.] \underline{Kind of} a coarse kind of test \\

\vspace{-0.5cm}

\item [14.] So we could pour concrete and support it with \hl{{\underline{a}}} a nice firm \underline{four} by four posts\\

\vspace{-0.5cm}

\item [15.] But uh \underline{\smash{I'm afraid {\hl{I'm}}}} I'm probably in the minority \\

\vspace{-0.5cm}

\item [16.] \hl{{\underline{\smash{Same thing}}}} same thing that \hl{the} her kids had\\

\vspace{-0.5cm}

\item [17.] \hl{Did {\underline{\smash{you}}} you framed it in} uh \hl{{\underline{on}} on} you framed in new square footage\\

\vspace{-0.5cm}
\item [18.] \hl{{\underline {And and}}} and there needs to be a line drawn somewhere \hl{at} reasonable and proper \\

\vspace{-0.5cm}
\item [19.] $\dots$~I think there's a couple \underline{of} levels of tests in terms
\hl{{\underline {of}}} of drugs \\

\vspace{-0.5cm}
\item [20.] See \hl{{\underline{\smash{they have}}}} uh we have \hl{two the} both c spans here 

\end{enumerate}

In some cases where repetitions are fluent, the model has incorrectly detected the first occurence of the word as disfluency (e.g. 13, 14, 15, 19). Moreover, when there is a long distance between reparandum and repair words (e.g. 11, 12), the model usually fails to detect the reparanda. In some sentences, the model is also unable to detect the disfluent words which result in ungrammatical sentences (e.g. 16, 17, 18, 20). In these examples, the undetected disfluencies ``the'', ``did'', ``at'' and ``two the'' cause the residual sentence to be ungrammatical. 

We also discuss the types of disfluency captured by the ACNN model, but not by the baseline CNN. In the following examples, the ACNN predictions (\underline{underlined} words) are the same as the ground truth disfluency labels (\hl{highlighted} words). The \textbf{bolded} words indicate the CNN prediction of disfluencies.

\begin{enumerate}
\item [21.] Uh well \hl{{\underline{\smash{\textbf{I} actually my dad's}}}} my dad's almost ninety \\

\vspace{-0.5cm}
\item [22.] \hl{{\underline{Not a man}}} not a repair man but just a friend \\ 

\vspace{-0.5cm}
\item [23.] \hl{{\underline{\smash{we\textbf{'re} from a county}}}} we're from the county they marched in  \\

\vspace{-0.5cm}

\item [24.] \hl{{\underline{Now let's}}} now we're done \\
\vspace{-0.5cm}

\item [25.] And \hl{{\underline{\smash{they've}}}} most of them have been pretty good   \\

\vspace{-0.5cm}

\item [26.] I do \hl{{\underline{as far as uh as far as}}} uh as far as immigration as a whole goes  \\

\vspace{-0.5cm}

\item [27.] No need \hl{{\underline{to use this}}} to play around with this space stuff anymore \\

\vspace{-0.5cm}

\item [28.] We couldn't survive \hl{{\underline{\smash{\textbf{in a} in a juror}}}} in a trial system without a jury \\

\vspace{-0.5cm}

\item [29.] You stay \hl{{\underline{\smash{within your}}}} uh within your means \\

\vspace{-0.5cm}

\item [30.] So \hl{{\underline{\smash{\textbf{we're} we're part}}}} we're actually part of MIT

\end{enumerate}

The ACNN model has a generally better performance in detecting ``rough copies'' which are important indicator of \emph{repetition} (e.g. 21, 29), \emph{correction} (e.g. 22, 23, 24, 25, 27), and \emph{stutter-like} (e.g. 26, 28, 30) disfluencies.

\section{Conclusion} \label{sec:6}
This paper presents a simple new model for disfluency detection in spontaneous speech transcripts.  It relies on a new auto-correlational kernel that is designed to detect the ``rough copy'' dependencies that are characteristic of speech disfluencies, and combines it with conventional convolutional kernels to form an auto-correlational neural network (ACNN). We show experimentally that using the ACNN model improves over a CNN baseline on disfluency detection task, indicating that the auto-correlational kernel can in fact detect the “rough copy” dependencies between words in disfluencies. The addition of the auto-correlational kernel permits a fairly conventional architecture to achieve near state-of-the-art results without complex hand-crafted features or external information sources. 

We expect that the performance of the ACNN model can be further improved in future by using more complex similarity functions and by incorporating similar kinds of external information (e.g. prosody) used in other disfluency models. In future work, we also intend to investigate other applications of the auto-correlational kernel. The auto-correlational layer is a generic neural network layer, so it can be used as a component of other architectures, such as RNNs. It might also be useful in very different applications such as image processing.

\section*{Acknowledgments}

We would like to thank the anonymous reviewers for their insightful comments and suggestions. This research was supported by a Google award through the Natural Language Understanding Focused Program, and under the Australian Research Council’s Discovery Projects funding scheme (project number DP160102156).

\bibliography{emnlp2018}
\bibliographystyle{acl_natbib_nourl}

\end{document}